%% file: root.tex
\documentclass[letterpaper, 10 pt, journal, twoside]{ieeetran}  
\usepackage{cite}
\usepackage{amsmath,amssymb,amsfonts}
\usepackage{algorithmic}
\usepackage{graphicx}
\usepackage{textcomp}
\usepackage{epsfig} 
\usepackage{mathptmx} 
\usepackage{times} 
\usepackage[utf8]{inputenc}
\usepackage{float}
\usepackage{epstopdf}
\usepackage{textcomp}
\usepackage{multirow}
\usepackage{enumitem}
\usepackage{color,soul}
\usepackage{subcaption}
\usepackage{caption,setspace}
\usepackage{comment}
\usepackage[normalem]{ulem}
\includecomment{track}
\newcommand{\movefrom}[1]{}
\newcommand{\del}[1]{}

\begin{track}
    \renewcommand{\del}[1]{\textcolor{red}{\sout{#1}}}
    
\end{track}

\author{Shrey Pareek$^{1}$, Harris J Nisar$^{2}$, and Thenkurussi Kesavadas$^{3}$%

\thanks{This work was supported by the National Science Foundation (NSF) under Grant No. 1502339.} 

\thanks{$^{1}$Shrey Pareek* is with the Data Science Department at Target, Minneapolis, MN, USA}

\thanks{$^{2}$Harris J Nisar and Thenkurussi Kesavadas are with the Department of Industrial and Enterprise Systems Engineering, University of Illinois, Urbana-Champaign, IL, USA.}

\thanks{$^{3}$Thenkurussi Kesavadas is the Vice President for Research and Economic Development
Executive Council at State University of New York at Albany, NY, USA

{\tt\footnotesize *mail@shreypareek.com}}%

\thanks{Digital Object Identifier (DOI): 10.XXXX.2020.XXXX}
}

\title{AR3n: A Reinforcement Learning-based Assist-\\As-Needed Controller for Robotic Rehabilitation}

\begin{document}

\maketitle

\begin{abstract}
In this paper, we present AR3n (pronounced as Aaron), an assist-as-needed (AAN) controller that utilizes reinforcement learning to supply adaptive assistance during a robot assisted handwriting rehabilitation task. Unlike previous AAN controllers, our method does not rely on patient specific controller parameters or physical models. We propose the use of a virtual patient model to generalize AR3n across multiple subjects. The system modulates robotic assistance in realtime based on a subject's tracking error, while minimizing the amount of robotic assistance. The controller is experimentally validated through a set of simulations and human subject experiments. Finally, a comparative study with a traditional rule-based controller is conducted to analyze differences in assistance mechanisms of the two controllers. 
\end{abstract}

\begin{IEEEkeywords}
Rehabilitation Robotics, Deep Learning in Robotics and Automation, Reinforcement Learning. 
\end{IEEEkeywords}

\input{contents}

\bibliographystyle{IEEEtran}
\bibliography{references}
\addtolength{\textheight}{-12cm} 
\end{document}

%% file: contents.tex
\section{INTRODUCTION} \label{sec:intro}

\IEEEPARstart{R}{ecent} years have seen the advent of robot-based rehabilitation systems as a reliable tool for home-based stroke therapy \cite{manjunatha2021upper}. These systems can provide autonomous robotic assistance to a patient as they perform prescribed therapy tasks using a simulation system. Robotic assistance is usually based on a set of rules that govern \textit{when} and \textit{how} to provide assistance to a patient. The choice of this assistance mechanism is a non-trivial task and serves as a crucial factor towards the success of robotic therapy  \cite{schmidt2008motor}. Inadequate assistance may render a task too difficult for the patient, inducing anxiety and forcing them to quit the rehabilitative task early \cite{nakamura2014concept}. Conversely, excessive assistance can lead to over-reliance on the robot \cite{schmidt2008motor}.

Assist-As-Needed (AAN) controllers \cite{maaref2016bicycle} provide adequate assistance by dynamically adjusting robotic assistance levels based on patient performance. In other words, as the user's performance improves, robotic assistance is reduced; and vice-versa. 

The simplest AAN controller is a rule-based error reduction (ER) \cite{manjunatha2021upper} mechanism. ER describes a strategy that minimizes tracking error in a path following task. Assistance is supplied based on two manually tuned parameters viz. robotic gain and maximal allowable error threshold. This strategy describes a force field at the boundary of the error threshold that restricts free subject motions to within the boundaries of this zone. If the subject deviates outside this zone, the robotic device provides a corrective force and guides them back inside this zone. However, the selection of robotic gain and zone size is not automatized and needs to be determined by a therapist. The lack of automation may lead to over-reliance on robotic assistance and can limit the rehabilitation outcome  \cite{schmidt2008motor}. 

Several studies have proposed AAN methodologies that circumvent the above over-reliance issue by automating and adapting robotic impedance based on subject performance. They implemented non-trivial mechanisms to obtain a model of subject performance. These models can be broadly categorized as physical models \cite{emken2005robot, crespo2008haptic, maaref2016bicycle, shahbazi2018position} and physiological signal-based models \cite{george2012combining, Esfahani2011, pareek2019myo}. Such models are generally patient specific and cannot be generalized to larger populations. In this paper, we propose a Reinforcement Learning (RL)-based generalizable adaptive AAN controller that automatically adjusts robotic assistance based on a subject's performance. 

The paper is organized as follows: Section \ref{sec:lit_review} surveys existing AAN controllers. Section \ref{sec:methods}  provides a description of the key components of the proposed system and presents experimental evaluations under various settings. Results are presented in Section \ref{sec:res}  and we conclude in Section \ref{sec:conclusion}.

\section{LITERATURE REVIEW} \label{sec:lit_review}

\subsection{Assist-As-Needed Controllers}
According to the guidance hypothesis \cite{schmidt2008motor}, humans demonstrate a tendency of over-reliance on external assistance, which may inhibit motor recovery. This has led to the inception of AAN controllers that adapt degree of robotic assistance based on subject performance.

Reinkensmeyer's group \cite{emken2005robot} proposed the use of patient-specific computational learning models that predict how subjects adjust their motor behavior in the presence of varying external forces. 
Crespo et al. \cite{crespo2008haptic} developed a wheelchair steering AAN controller. However, the approach requires 25-40 trials per subject to develop a subject-specific assistance model. Maaref et al. \cite{maaref2016bicycle} proposed a task difficulty model to estimate the difference between a patient's motor skills and task difficulty to toggle robotic assistance. However, the algorithm relies on learning the \textit{ideal} impedance and/or position tracking behavior exhibited by an expert through multiple demonstrations while conducting a therapy task. 
\cite{shahbazi2018position} used the concept of passivity to describe the maximum amount of assistive force that can be \textit{safely} absorbed by a patient's arms. Robotic impedance can then be modulated within this safety limit to deliver stable and adaptive assistance. Estimation of maximal assistive force is non-trivial and requires periodic calibrations sessions with the patient. In our previous work \cite{pareek2019iart}, we proposed iART, that uses demonstrations from an expert to mimic and recreate their assistance behavior. 

Brain Computer Interface (BCI) and Surface Electromyography (sEMG) sensors can be used to develop physiological models instead of the patient specific physical models described above. These methods use BCI \cite{george2012combining, Esfahani2011} and sEMG \cite{pareek2019myo} signals to adapt robotic assistance based on a patient's mental engagement and amount of physical effort applied by them, respectively. These sensors need a considerable amount of time to set up and usually require the assistance of another person in doing so. This limits their feasibility as a home-based rehabilitative tool. \cite{manjunatha2021upper} provides a scoping review of adaptive assistance techniques for rehabilitation robotics. In this paper, we propose the use of a RL-based controller that circumvents the challenges associated with deriving complex subject-specific physical models  and the feasibility issues of external sensor-based systems.

\begin{figure*}[!t]
	\centering
	\includegraphics[ width=\linewidth,  trim={0 0cm 0 0cm},clip]{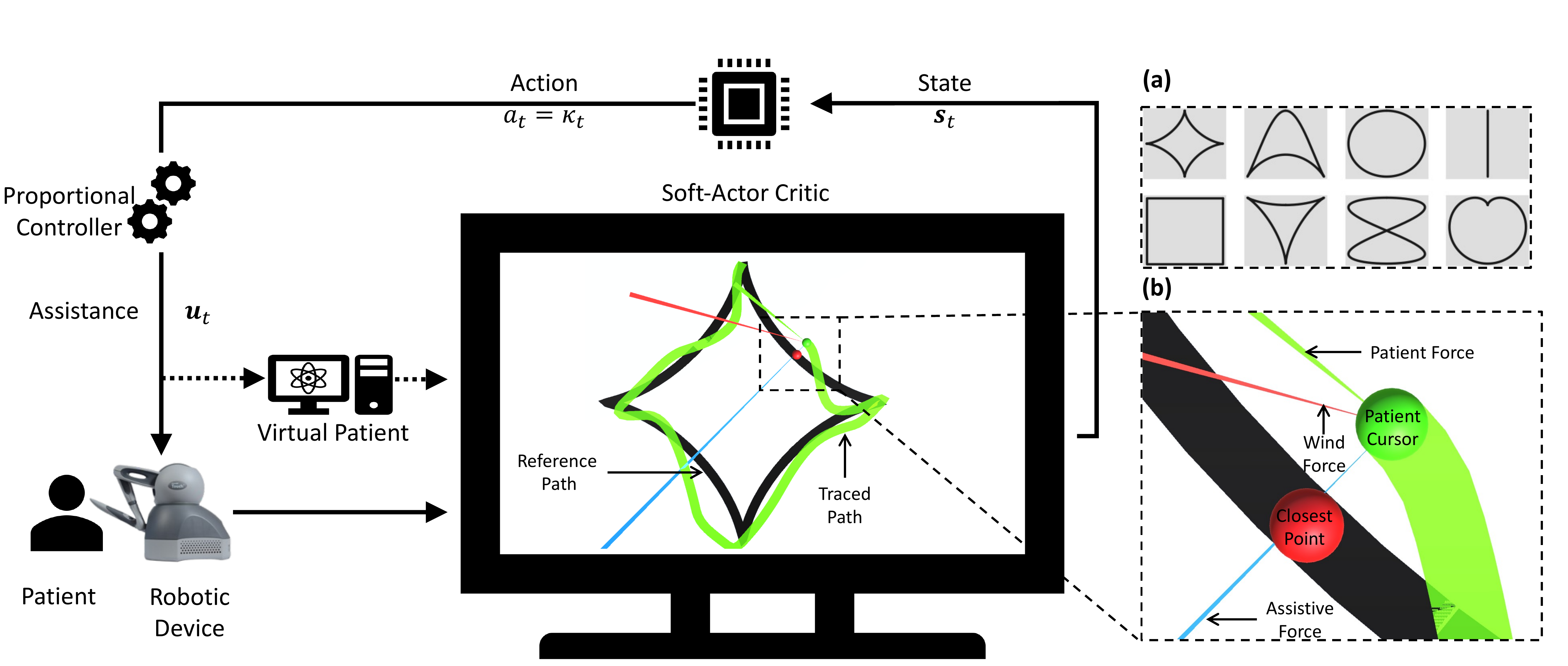}
	\caption{Schematic representation of AR3n. The three key components viz. simulation environment, SAC RL module, and robot motor task are shown. During training, assistance is applied to the virtual patient (dotted arrows). Once trained, assistance is directly supplied to a human subject through a robot. $s_t$ denotes current state at time $t$. $a_t$ represents agent action which in this case is the controller gain $k_t$. $u_t$ represents robotic assistance supplied through controller action. Inset (a): Reference trajectories used for training (top) and testing (bottom). Inset (b): Virtual patient force model used to simulate multiple patients while training the RL agent.}
	\vspace{-10pt}
	\label{fig:robot}
\end{figure*}

\subsection{Reinforcement Learning-based AAN Controllers}

RL describes a set of learning mechanisms that learn an optimal mapping between situations and actions so as to maximize a numerical reward signal \cite{sutton2018reinforcement}. An RL agent derives the optimal policy for a given Markov Decision Process (MDP) based on data acquired through exploration and experience. RL has gained popularity in recent years across various robotics-based domains. However, very few endeavours have been made towards the development of RL-based robotic AAN controllers. Obayashi et al. \cite{obayashi2014assist} developed one of the earliest RL-based AAN controllers. Using dart-throwing as a case study, the authors proposed a user-adaptive robotic trainer that aims at maximizing the score in a game of darts while minimizing physical robotic assistance. \cite{hamaya2016learning} demonstrated the use of model-based RL in conjunction with sEMG for formulating effective assistive strategies for exoskeleton-based systems. 

Inverse RL (IRL) is another strategy wherein a desired policy is derived from expert demonstrations. Scobee et al. \cite{scobee2018haptic} demonstrated the use of IRL to provide haptic assistance. \cite{reddy2018shared} proposed a human-in-the-loop RL algorithm to demonstrate shared autonomy through the Lunar Lander video game and a real quadcopter. Luo et al. \cite{luo2021reinforcement} used the Proximal Policy Optimization (PPO) RL algorithm to provide assistance via an exoskeleton during squatting motion. Their approach relies on a task specific reference motion demonstration for learning. The reliance on expert human demonstrations serves as a limitation to these IRL-based studies. The method proposed in this paper aims at eliminating the reliance on human participation/demonstrations during the RL model training phase. 

More recently, \cite{zhangadaptive} used an actor-critic RL algorithm to modify robotic impedance for ankle mobilization. Impedance is adjusted to minimize tracking error while a control objective determines the amount of assistance to be supplied to the subject. The RL agents learns an optimal policy that modifies robotic impedance to achieve the desired control objective. The results were reported on a predefined sinusoidal trajectory and showed greater improvements in learning when compared with a conventional AAN controller. The above methods \cite{obayashi2014assist, hamaya2016learning, zhangadaptive} yield subject-specific controllers based on online RL training. The method proposed here uses Soft Actor Critic (SAC)-based RL \cite{haarnoja2018soft} to generalize assistance mechanism across multiple patient behaviors. 

We introduce AR3n\footnote{pronounced as Aaron.}, an \textbf{A}ssitive \textbf{R}obotic \textbf{R}ehabilitation system based on \textbf{R}einforcement Lear\textbf{N}ing. AR3n uses RL to dynamically adjust robotic assistance  and does not require patient-specific physical models or physiological sensors to estimate the same. We achieve this by simulating a plethora of patient behaviors through a virtual patient model and training a RL-based assistant to generalize across these behaviors. First use human-subject studies are conducted to test the assistance behavior of the virtual patient trained model on healthy subjects. The development of a virtual patient model and an RL-based AAN controller are the key contributions of this work. To the best of the authors' knowledge, this is the first study that uses a simulation-based upper-limb RL-AAN controller. A demo video for AR3n may be found here \footnote{https://youtu.be/hTVjd7uzMz8}.

\section{METHODS} \label{sec:methods}

AR3n comprises of three key components (see Fig. \ref{fig:robot}), viz.  (i) simulation environment: with which the RL agent interacts, (ii) RL module: that uses SAC to learn and predict robotic assistance, and (iii) robotic motor task: that uses the trained RL agent from (ii) to deliver realtime adaptive robotic assistance.

\subsection{Motor Task} \label{sec:motor}

In this study, we have used the example of handwriting rehabilitation. A subject uses a robotic end effector to control the position of a virtual pen in a handwriting simulation environment (see Fig. \ref{fig:robot}). The robotic device provides kinesthetic assistance to the patient based on prescribed control mechanisms such as ER or AR3n. The writing simulation environment was developed using  Unity3D\footnote{https://unity.com} and consists of a virtual environment wherein, a reference trajectory to be followed by the patient can be chosen from Fig. \ref{fig:robot} Inset-(a). 

A robotic device assists the user along the trajectory based on a proportional (P) controller. Traditionally, the gain of a P-controller is chosen by a therapist. In this paper, we propose a methodology to adapt this parameter automatically based on the subject's performance. A 6 degrees-of-freedom (6 revolute joints: 3 actuated and 3 passive) Geomagic\textsuperscript{\textregistered} Touch\textsuperscript{TM} was used in this study to provide kinesthetic feedback to the user at a sampling rate of 1000 Hz. The actuated joints can provide force feedback up to $3.3N$ to the user.

\subsection{Reinforcement Learning Module}

This module comprises of a SAC-based agent that interacts with a simulation environment to learn the optimal assistance policy. The agent's goal is to derive a mapping between state ($\mathbf{s}$) and action ($a$) that maximizes the cumulative return $V$ from the current reward $r$ \cite{sutton2018reinforcement}. This is usually achieved through the means of a simulation environment within which an agent must be able to take actions that affect the state of the environment. To formalize the problem being addressed in this paper, we first describe a handwriting simulation task which has been modelled as an MDP.

\subsubsection{Training Environment}

In this paper, we use a robot-assisted hand writing task as the case study. The patient's goal is to track a reference path using the robotic device. However, the patient's motor deficits may prevent them from achieving low tracking error. The RL agent serves the role of a therapist and decides when and how to assist a patient based on their performance (see Fig. \ref{fig:robot}). The agent learns this assistance behavior by interacting with the environment for data acquisition through exploration and experience. The need for large quantities of data for effective learning prevents the use of real subject-robot interaction (see Section \ref{sec:motor}) while training the RL agent. As a result, we simulate the handwriting environment as well as the patient. The training task is designed as an episodic task, wherein each episode involves a virtual patient tracking a reference shape chosen randomly from the top row of Fig. \ref{fig:robot} Inset-(a). An episode in RL refers to a sequence of states, actions, and rewards with a terminal state. Terminal state in this scenario refers to reaching the end point of the reference trajectory to be tracked. 

\subsubsection{Virtual Patient Force Model} \label{sec:patient_model}

\begin{figure}[t!]
	\centering
	\includegraphics[ width=\linewidth]{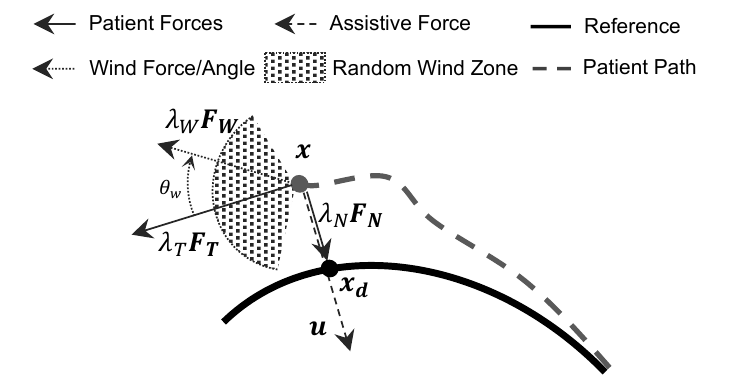}
	\caption{Virtual patient force model. Three types of patient forces are represented viz. a tangential force ($\mathbf{F_T}$), a normal force ($\mathbf{F_N}$), and a wind force ($\mathbf{F_W}$) that acts at an angle $\theta_w$. $x$ and $x_d$ denote the current patient position and the desired position, respectively. $\mathbf{u}$ is the assistive force used to correct the patient's trajectory.}
	\vspace{-10pt}
	\label{fig:patient_forces}
\end{figure}

We present a virtual patient model (see Fig. \ref{fig:patient_forces}) that enables us to simulate numerous patient behaviors and allows the RL agent to train and generalize across these behaviors. This circumvents the requirement of human subjects during training. 

The patient is simulated as a combination of three different types of forces, viz. a tangential force ($\mathbf{F_T}$), a normal force ($\mathbf{F_N}$) and a wind force ($\mathbf{F_W}$). $\mathbf{F_T}$ refers to a force tangential to the reference path that enables the patient to travel along the path. $\mathbf{F_N}$ is normal to $\mathbf{F_T}$ and describes the ability of the patient to minimize their tracking error by \textit{pulling} them towards the reference path. $\mathbf{F_T}$ and $\mathbf{F_N}$ collectively describe the ability of a patient to track the reference path in the absence of any motor impairments. Motor impairments are simulated as a random wind force ($\mathbf{F_W}$) acting in a random direction ($\mathbf{\theta_W}$) along the path. Total resultant patient forces are described as:

\begin{eqnarray}
    \mathbf{F_{P1}} &= \lambda_T \mathbf{F_T} + \lambda_N \mathbf{F_N} + \lambda_W \mathbf{F_W} \label{eq:patient_forces}
\end{eqnarray}

where, $\lambda_*$ is a scaling factor that decides the \textit{strength} of force $\mathbf{F_*}$. 
We chose $\lambda_T=1$ and $\lambda_N=0.4$ experimentally as they enabled low error trajectory tracking similar to that expected from a healthy person. $\lambda_T=1$ and $\lambda_N=0.4$ served as baselines around which all other forces were scaled experimentally such that they yielded realistic-looking trajectories. Realistic here refers to a heuristic wherein we visually verified that generated trajectories \textit{appeared} similar to those that would be reasonably exhibited by patients. 

$\lambda_W$ is randomly set to a value between $1.8$ and $2.2$. $\lambda_W$ is set higher than $\lambda_N$ to ensure deviations from the path. The wind angle ($\theta_W$) is randomly chosen between $-\pi/3$ and $+\pi/3$\footnote{$\pm \pi/3 $ was chosen as the wind angle range since values larger than this did not yield realistic looking trajectories.} (hatched sector of influence in Fig. \ref{fig:patient_forces}). Wind direction and magnitude ($\lambda_W\in[1.8,2.2]$) are varied every $0.75s$ to $1.5s$ during simulation runs. This high variability in terms of wind direction, magnitude, and variation frequency enables us to simulate multiple patient behaviors on which to train the RL agent. Since these parameters are not explicitly supplied to the agent, the proposed methodology operates as model-free RL.

\subsubsection{Formulating the Reinforcement Learning Problem}\label{sec:formulate-rl}

Formulating the RL problem requires formalizing it as an MDP. A well posed MDP consists of a tuple of states ($\mathbf{s}$), actions ($a$) and reward ($r$). The state at any time-step $t$ is given as: $\mathbf{s}_t=[e_t, e_{t-1},..., e_{t-n-1}]\in \mathbb{R}^{n\times1}$.

Where, $e_t$ refers to the perpendicular distance between the current patient position its orthogonally projected closest point ($x_d$) on the reference path at time-step $t$. Since only tracking error is used to describe the state of the system, the behavior of RL agent is reference trajectory agnostic and does not require retraining for generalization to reference trajectories not used during training. We also provide the tracking error at the previous $n-1$ steps as the state. This takes into account historic performance of the patient in addition to their instantaneous behavior. We set $n=25$ in this study which is equivalent to $0.5s$ of history (sampling rate of the simulation environment is $50 Hz$). 

The agent action is given by $a=\kappa\in[0,1]$, which is the gain of the proportional controller given as:

\begin{eqnarray}
    \mathbf{u}_t= \rho \kappa_{t}[\mathbf{x_d}_t-\mathbf{x}_t] \label{eq:ic}
\end{eqnarray}

where, $\mathbf{u}\in \mathbb{R}^{2\times1}$ is the assistive force being supplied to the patient. $\mathbf{x}\in \mathbb{R}^{2\times1}$ and $\mathbf{x_d}\in \mathbb{R}^{2\times1}$ denote the current cursor position and the desired point on the path. $\rho=3$ is a scaling factor to scale the gain predicted by the agent. This value ensures that
the maximum assistance is strong enough to assist the subject. 

The assistive force derived from (\ref{eq:ic}) acts on the existing patient forces described by $\mathbf{F_{P1}}$ in the same direction as $F_N$ to give the net patient force $\mathbf{F_{P}}$. In case of the actual motor task (Section \ref{sec:motor}), the assistive force is converted to torque values applied at the joints of the robotic device.
\begin{eqnarray}
    \mathbf{F_P} = \mathbf{F_{P1}} + \mathbf{u} \label{eq:total_force}
\end{eqnarray}

The instantaneous reward $r$ is a continuous function of tracking error and amount of assistive force applied. The expected cumulative return $V$ is the discounted sum of future rewards given by:

\begin{subequations}
\label{eq:reward_value}
\begin{align}
    r_t &= -\alpha \hat{e}_t - \beta \hat{u}_t -\delta \dot{\kappa}_{t}^2  \label{eq:reward}\\
    V_t &= \sum_{k=0}^{\infty}\gamma^kr_{t+k+1} \label{eq:return}
\end{align}
\end{subequations}
where, $\hat{e} = \frac{1}{n}[\sum_{k=0}^{n}e_{t-k}]^2$ describes a quadratic penalty associated with the average tracking error over the past $n$ steps. $\hat{u} = \frac{1}{n}\sum_{k=0}^{n}\left\lVert\mathbf{u_{t-k}}\right\lVert_2$ is the average assistive force magnitude applied over this interval. The final term ($\dot{\kappa}$) in (\ref{eq:reward}) is a penalty associated with fast changes in values of the proportional gain $\kappa$ predicted by the SAC network. This promotes a smoother assistance behavior. In other words, the reward function penalizes tracking error while penalizing any assistive force being applied and/or changed by the agent.  

We conducted numerous training runs to arrive at these values. First, $\alpha$ was set to a unit reference value and $\beta$ and $\delta$ were varied from $0-1$. The weights for each term in (\ref{eq:reward}) were empirically determined as $\alpha=1,\beta=0.45,\delta=0.5$ as these values demonstrated more consistent training results with reasonable assistive performance. 

$\gamma\in[0,1]$ in (\ref{eq:return}) refers to a discounting factor which decides the importance of future v/s current rewards while calculating the expected returns of the current state-action pair. We give equal weightage to both and hence set $\gamma=0.5$. This ensures quick adaptation to the current state while preventing an overly short-sighted agent. Most RL applications use $\gamma>0.9$, to maximize cumulative reward over a larger time horizon. Since the reward window in this case is around $0.5s$ as described earlier in the section, $\gamma=0.5$ was a viable choice. 

\subsubsection{Soft-Actor-Critic Network} \label{sec:sac}

SAC \cite{haarnoja2018soft} is an off-policy method that uses a replay buffer to improve sample efficiency. This implies that network parameters are updated with experience collected from a different policy than the current one; allowing the algorithm to generalize over a larger state space without explicitly visiting them. Off-policy SAC was chosen by conducting a preliminary comparison of training performance with the PPO algorithm. We describe this analysis in Section \ref{sec:exp_training}. 

SAC is based on maximum entropy RL, with the following entropy augmented objective function: 

\begin{eqnarray}
    J(\pi) = \mathbb{E}\huge{[}\sum_{t}r(\mathbf{s}_t,a_t)-\chi\log(\pi(a_t|\mathbf{s}_t))\huge{]} \label{eq:objective}
\end{eqnarray}

An optimal policy $\pi^*$, maximizes the expected return and entropy ($\log$-term in (\ref{eq:objective})). The entropy term can be viewed as a trade off between exploration (maximize entropy) and exploitation (maximize return). The trade-off between the two is controlled by the non-negative temperature parameter $\chi \in[0,1]$. We set $\chi=0.5$ throughout the training process. 

A soft Q-function describes the critic, while a Gaussian policy function entails the actor. In other words, given a state $\mathbf{s}_t$ the actor chooses an action $a_t$ based on the stochastic policy $\pi_\phi$. Meanwhile, the critic estimates the expected returns of the current state-action pair using a soft Q-function $Q_\theta(\mathbf{s},a)$. As mentioned earlier, the action $a_t$ corresponds to gain $\kappa_t$, which modulates the assistive force $\mathbf{u}_t$ through (\ref{eq:total_force}). We refer the reader to  \cite{haarnoja2018soft} for more details on SAC. Both networks (actor and critic) use the same neural network architecture with three hidden layers and 32 units in each layer. The learning rate was set at $1e-5$ and a batch size of 128 was used. These hyperparameters were chosen as they yielded stable training with high average rewards during preliminary testing.

The SAC model was trained for 50K steps using the training environment and virtual patient model described above. Only the 4 shapes in the top row of Fig. \ref{fig:robot} Inset-(a) were used for training as the proposed formulation here is reference trajectory agnostic as described in Section \ref{sec:formulate-rl}. Once the model was trained, it was used for realtime inference. Training was performed on an  Intel Core i7 5820K Processor with a NVIDIA GeForce GTX 970 - 4GB graphics card and training times averaged around $20$ minutes.

\subsubsection{SAC Training Performance} \label{sec:exp_training}

\begin{figure}[!b]
	\centering
	\includegraphics[width=\linewidth, trim={0 .5cm 0 .5cm},clip]{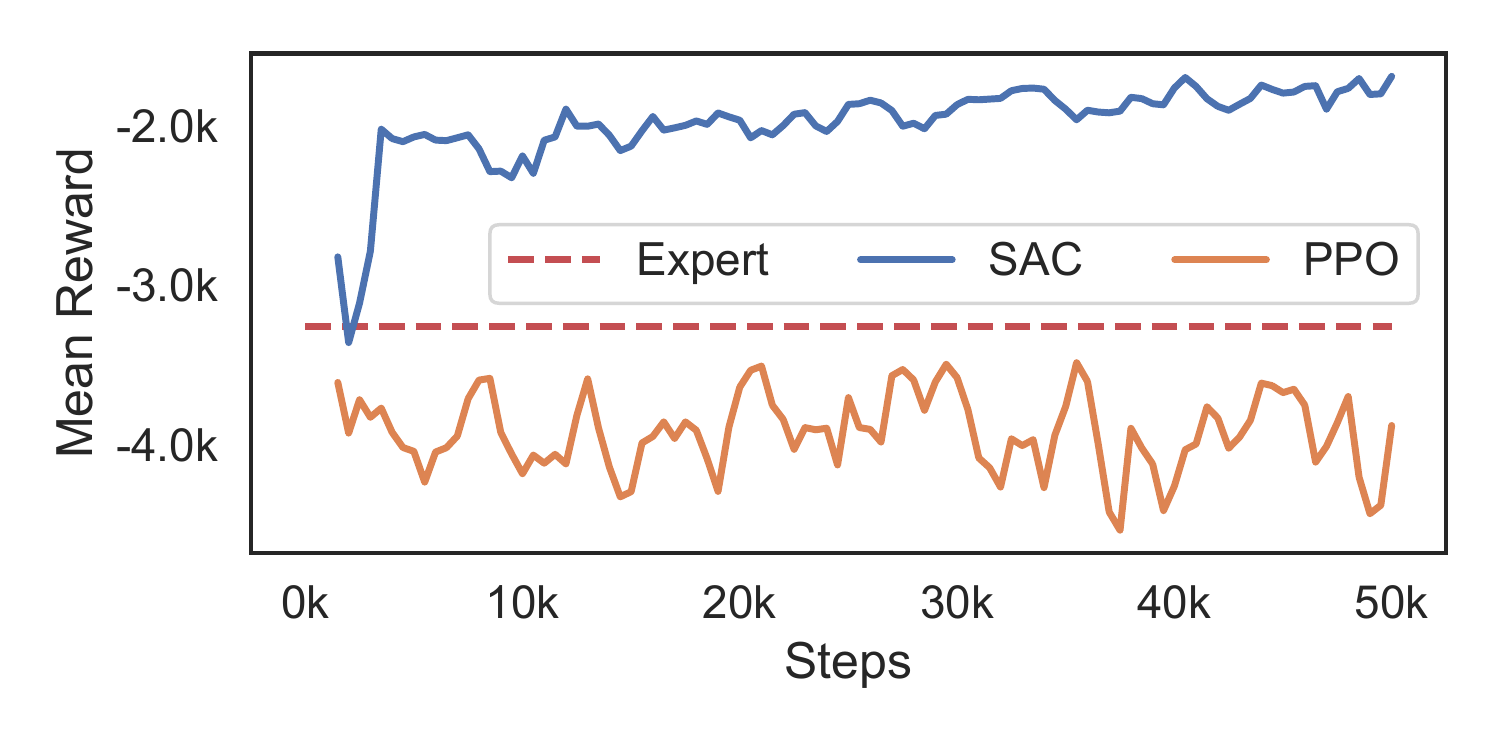}
	\caption{Mean reward v/s training steps for SAC and PPO. SAC outperforms PPO by a significant margin.}
	\vspace{-10pt}
	\label{fig:reward_steps}
\end{figure}

We conduced a pilot experiment to substantiate the choice of SAC for this study. We compared SAC's training performance in terms of average reward per episode with the PPO on-policy algorithm. PPO is a widely used policy gradient RL method \cite{sutton2018reinforcement} that finds applications in continuous action tasks. 
 
Fig. \ref{fig:reward_steps} shows average reward per episode v/s training steps for SAC and PPO. The constant dotted line demonstrates performance of an expert human. One of the authors served as the expert and toggled assistance on and off as the virtual patient model simulated 10 episodes. It should be noted that the expert reward here is only presented as reference and not for comparative analysis. 

Unsurprisingly, SAC outperformed PPO by a large margin and demonstrated very fast learning in terms maximizing the cumulative reward. This fast learning is attributed to the temporal difference learning methodology used by SAC. The superior performance of SAC when compared with PPO is in agreement with other studies \cite{haarnoja2018soft}. These results affirm the choice of SAC as a valid RL algorithm for this study and discard we PPO from further analysis. 

\subsection{Experimental Evaluation}
We designed experiments aimed at evaluating AR3n's ability to (i) conduct realtime inference in a simulated environment, and (ii) verify its ability to provide realtime assistance and induce motor learning with human subjects. We also compared differences between assistance mechanisms of AR3n and a traditional ER-based controller.

\subsubsection{Simulated Testing} \label{sec:exp_sim}
We evaluated AR3n in terms of delivering reliable online assistance with the ER assistance mechanisms. ER refers to the error reduction assistance mechanism described in Section \ref{sec:intro}. In this experiment, we used the virtual patient to test differences between the two assistance mechanisms. 
Both methods were used to modulate assistance in realtime for 50 virtual episodes. The same random seed was used for both cases. This enabled us to compare the assistance behavior of the two methods, subject to the exact same initial conditions. 

\begin{figure}[!b]
	\centering
	\includegraphics[ width=\linewidth]{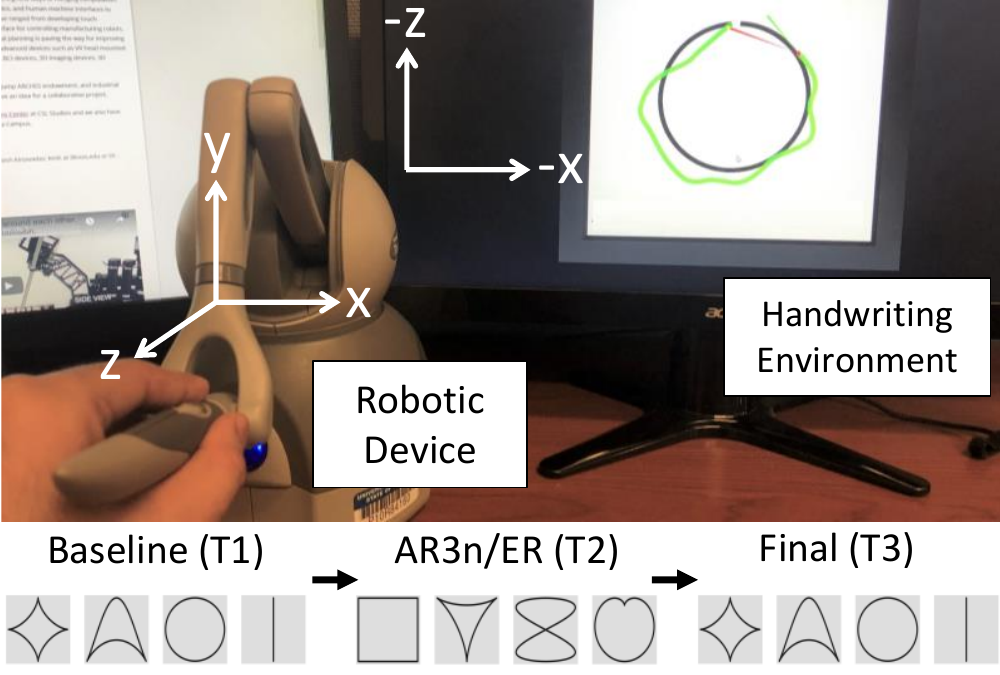}
	\caption{Experimental setup and human subject study design.}
	\vspace{-10pt}
	\label{fig:experiment}
\end{figure}

\subsubsection{Human Subject Study} \label{sec:exp_human}
Next, we conducted a first-use human subject study to (i) verify AR3n's ability to use the assistance behavior learnt using the virtual patient model to deliver realtime assistance to human subjects; and (ii) study differences in AR3n and ER as rehabilitative tools. Eight healthy subjects (5 males; 3 females; average age 26 years; range 19-33 years) were recruited for a single session approved by the University of Illinois at Urbana-Champaign's Institutional Review Board (IRB \#15990). 

The experimental setup is shown in Fig. \ref{fig:experiment} and involved the subject using a robotic device for the trajectory tracking task described in Section \ref{sec:motor}. In order to increase the task difficulty and simulate motor impairment, the subjects were required to use their non-dominant arm and the robot motions were mirrored in the horizontal ($x$) and transverse ($z$) direction (see Fig. \ref{fig:experiment}). In other words, if the robot end-effector was moved to the right, the onscreen cursor would move to the left, and vice-versa. A similar reversal was implemented in the $z-$direction. 

Each subject participated in three trials (Fig. \ref{fig:experiment}). A baseline trial (T1) followed by a training trial (T2), and a final post training trial (T3). Each trial involved the subject executing the four shapes shown in Fig. \ref{fig:experiment}. This set of shapes was chosen as it contains both straight line and curved sections. The size of these shapes was scaled to be roughly equivalent to a standard A4 sized paper to stimulate larger movements of the subject's arm. All trials lasted around 2 minutes with 2 minute breaks between subsequent sessions. A brief acclimatization trial (T0) was conducted so that subjects could familiarize themselves with the system.

Baseline and final trials involved no robotic assistance and used the same 2D shapes. During the training session (T2), robotic assistance was provided either through a conventional ER-based AAN or the proposed AR3n controller. The maximal allowable error for ER was set as $0.3$. This means that robotic assistance was toggled on only when tracking error was greater than $0.3$. The gain for ER was set at $3$. 

The eight subjects were randomly assigned to the conventional ER ($E_k$, $k=1,..4$) controller or AR3n ($A_k$, $k=1,..4$). The subjects were unaware of the type of assistance being supplied. The goal of this experiment was to compare the change in tracking error from T1 to T3 among the two groups.  

\section{RESULTS AND DISCUSSIONS} \label{sec:res}

\subsection{Simulated Testing} \label{sec:res_sim}

\begin{figure}[!b]
	\centering
      \begin{subfigure}[b]{0.18\textwidth}
         \centering
         \includegraphics[width=\linewidth, trim={15mm 15mm 25mm 15mm},clip]{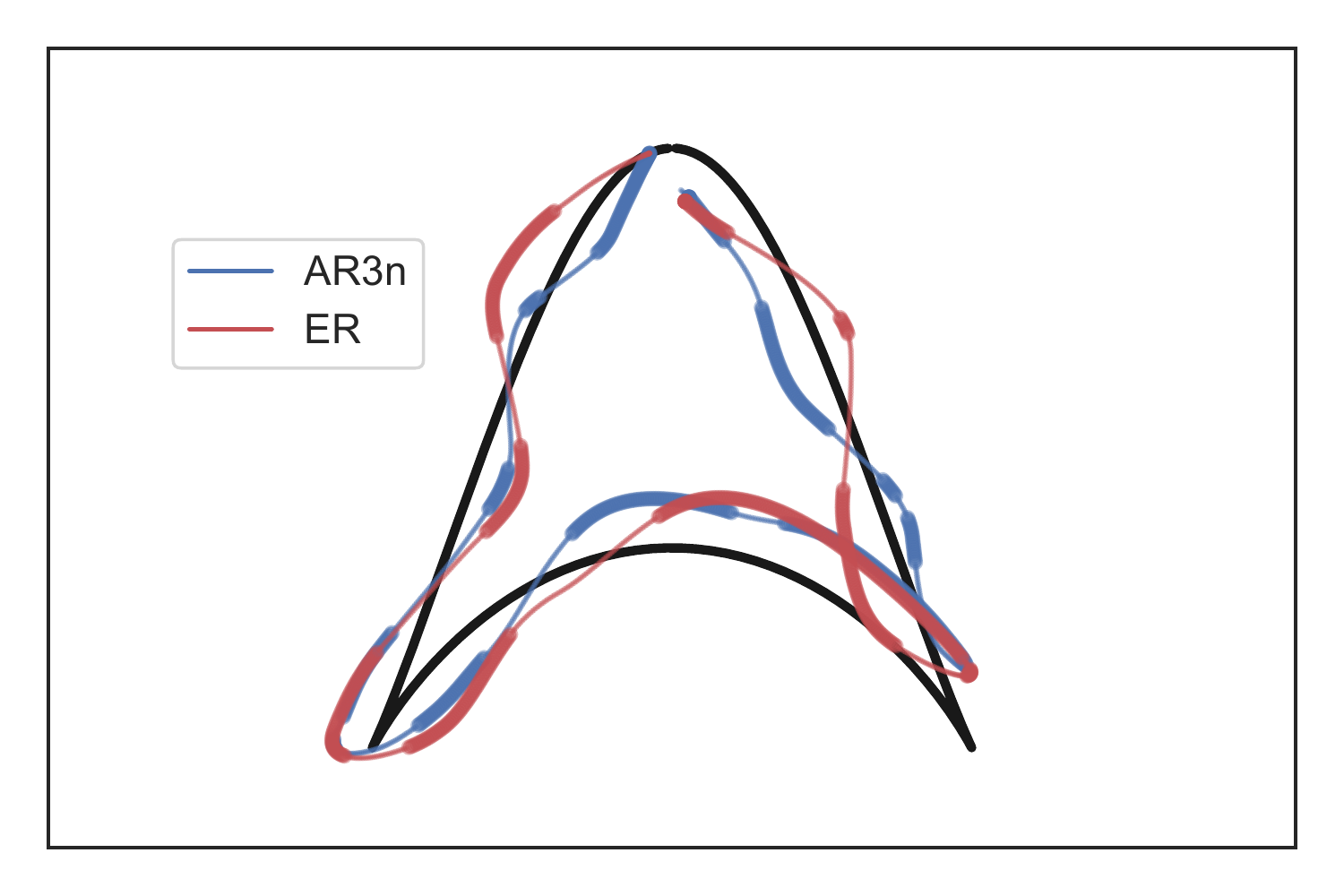}
     \end{subfigure}
     \begin{subfigure}[b]{0.3\textwidth}
         \centering
	\includegraphics[width=\linewidth, trim={0.5 .5cm 0 .5cm},clip]{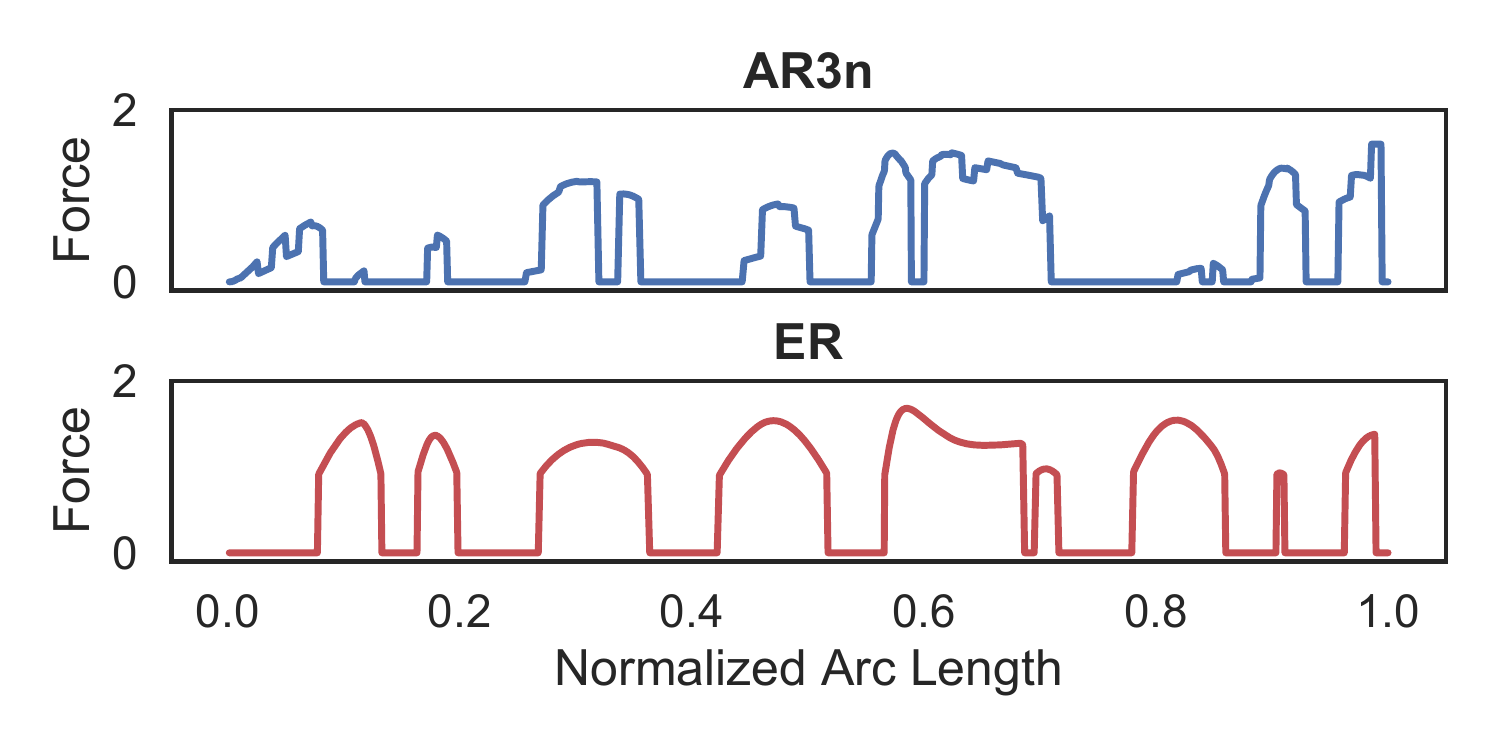}
     \end{subfigure}
     \hspace*{\fill}
 	\caption{(Left) Trajectories executed by virtual patient under AR3n (blue) and rule-based ER (red) controller. Size of dots denotes the amount of assistive force applied by the respective algorithms. (Right) Assistive force profiles under the two assistance settings. }
   \label{fig:assist_sim_compare}
\end{figure}

Fig. \ref{fig:assist_sim_compare}-Left shows tracking behavior executed by the virtual patient under AR3n and ER for the same trajectory and random seeds. Fig. \ref{fig:assist_sim_compare}-Right shows the corresponding assistive force modulation v/s arc length by the above assistance mechanisms. It can be observed that both mechanisms demonstrate similar assistance profiles in terms of \textit{when} assistance was provided, but differed \textit{how} assistance was provided. Owing to the rule-based nature of ER, assistance is provided in short busts similar to a step function. These \textit{bursts} can be smoothed using a mathematical function, however the key drawback of ER remains the use of a rule-based controller with manually determined thresholds. 

AR3n on the other hand modulates degree of assistance based on complex rules learnt using the virtual patient behavior. This smoother modulation of AR3n is attributed to the quadratic penalty associated with rapid gain switching ($\dot{\kappa}$ term in (\ref{eq:reward})). Eliminating this term would lead to an RL controller that learns a bang-bang optimal control policy, which is not suitable for assisted robotic rehabilitation. 

\begin{figure}[!t]
	\centering
	\includegraphics[ width=\linewidth]{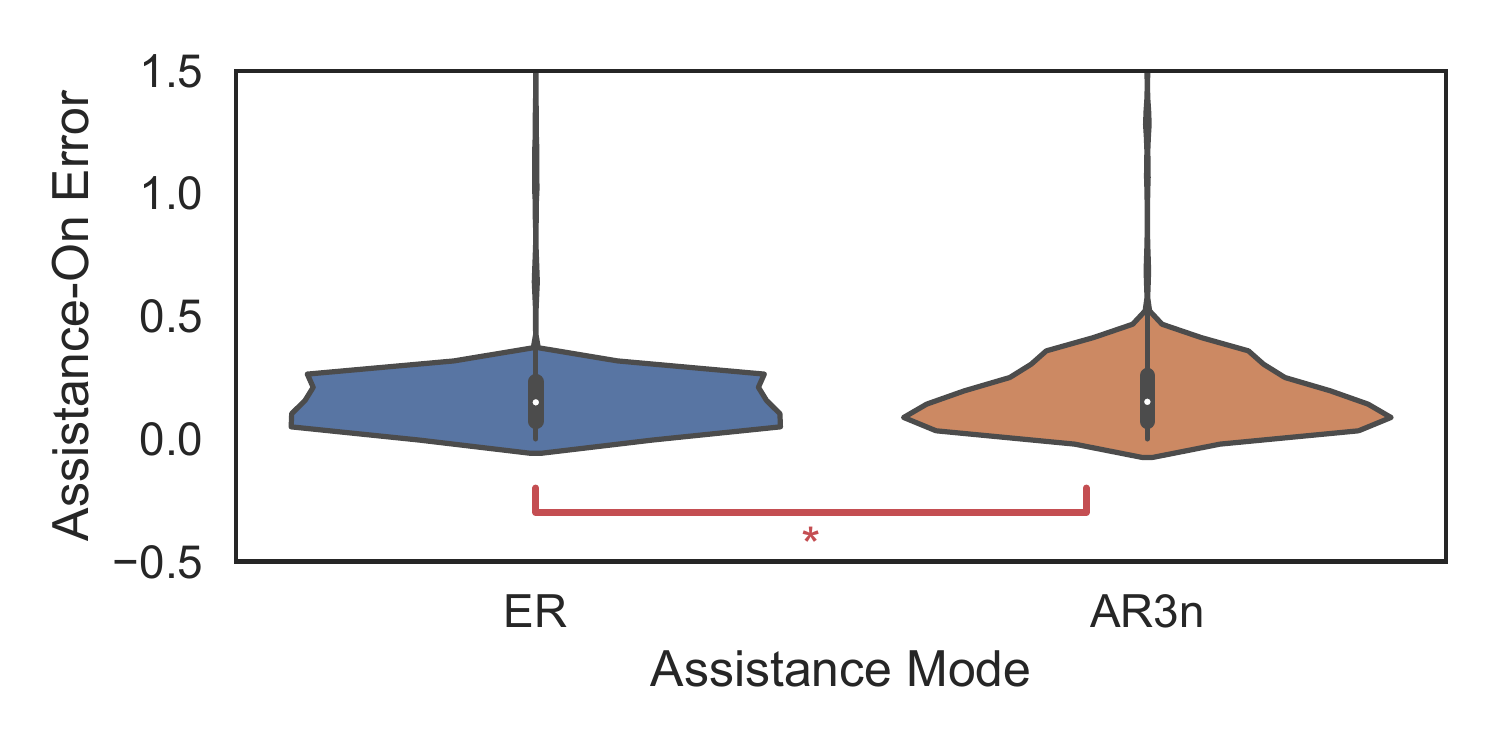}
	\caption{Violin plots for tracking error at which assistance was switched on using different assistance mechanisms. Asterisks denote significant differences.}
	\vspace{-10pt}
	\label{fig:violin_sim}
\end{figure}

We also compared the tracking error at which assistance  was switched on (gain$>0$) under AR3n and ER. These error distributions are shown as violin plots in Fig. \ref{fig:violin_sim}. Under ER, assistance-on error was concentrated around the error zone ($0.3$). This was expected, since ER only prevents the subject from exiting the force field and does not assist them by guiding them back towards the reference trajectory. As with ER, AR3n demonstrates a denser distribution around the error zone but with a wider spread overall. Assistance-on error for AR3n and ER shows significant differences ($t = -24.28$, $p = 10^{-20}<0.05$). 

\subsection{Human Subject Study} \label{sec:res_human}

 \begin{figure}[!b]
     \centering
    \includegraphics[width=\linewidth, trim={0 .5cm 0 .5cm},clip]{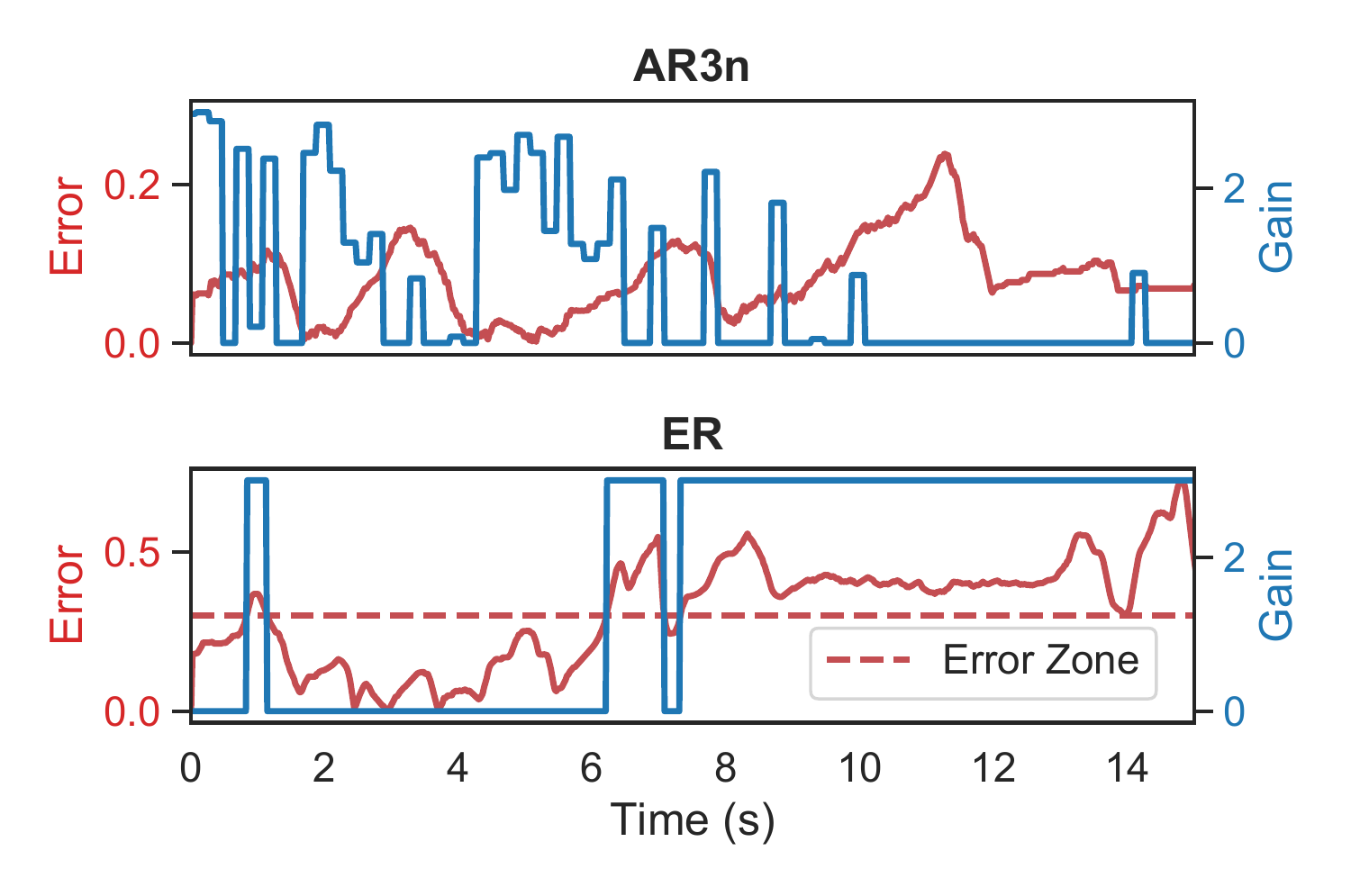}
     \caption{Variation of gain (blue) w.r.t. tracking error (red) under AR3n (top) and ER (bottom) for two test subjects.}
     \label{fig:gain_error_real}
     \vspace{-10pt}
 \end{figure}

Fig. \ref{fig:gain_error_real} shows gain variation (blue) under AR3n and ER w.r.t. tracking error (red) for two subjects. The dotted line demonstrates the size of error zone for ER. ER assists the subject only when the tracking error is higher than this threshold. AR3n on the other hand, does not follow a strict error-based rule while deciding how much to assist the subject. 

Under ER the subject tends to over-rely on the robotic assistance as is evident from the near continuous assistance provided from $6-14s$. Additionally, ER merely prevents the subject from deviating outside the error zone boundary, it does not assist them by guiding them back to the reference trajectory.

 \begin{figure}[!t]
	\centering
	\includegraphics[width=\linewidth, trim={0 .5cm 0 .5cm},clip]{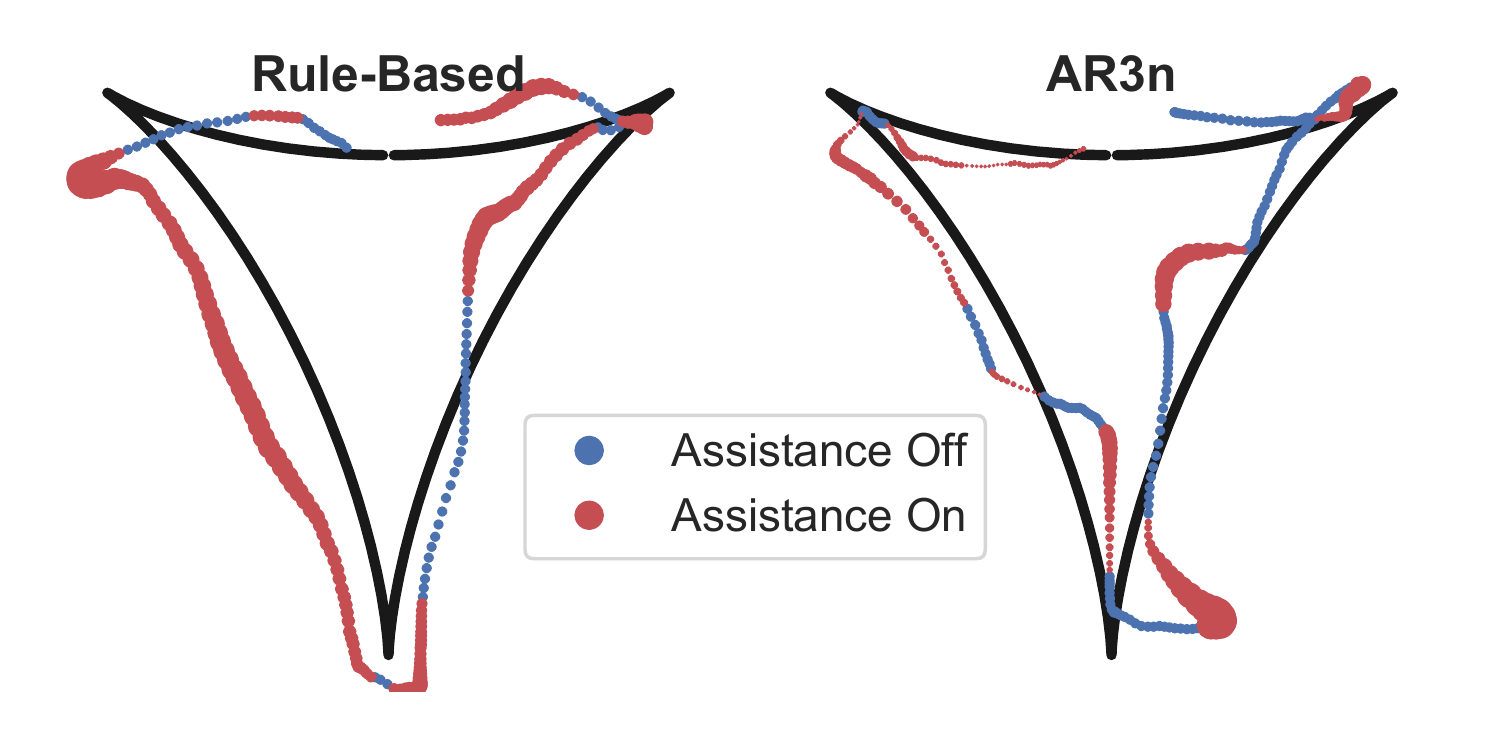}
	\caption{Tracking behavior demonstrated under rule-based ER and AR3n. Size of red dots denotes the amount of assistive force applied by the respective algorithms.}
	\vspace{-10pt}
	\label{fig:comparison}
\end{figure}

The over-reliance tendency under ER can be visualized in Fig. \ref{fig:comparison}-Left. The figure presents trajectories executed by a subject under ER.  
It can be observed that the subject tends to stay at the boundary of the force field, as at the boundary, the robotic device provides minimal assistance enabling the subject to correctly follow the trajectory with minimal effort and low tracking error. AR3n (\ref{fig:comparison}-Right) on the other hand guides the subject back to the reference trajectory and then switches off assistance. This is also evident from the reduction of tracking error from a large value to near-zero whenever assistance was switched on in Fig. \ref{fig:gain_error_real}.

\begin{figure}[!b]
	\centering
	\includegraphics[width=\linewidth, trim={0 .5cm 0 .5cm},clip]{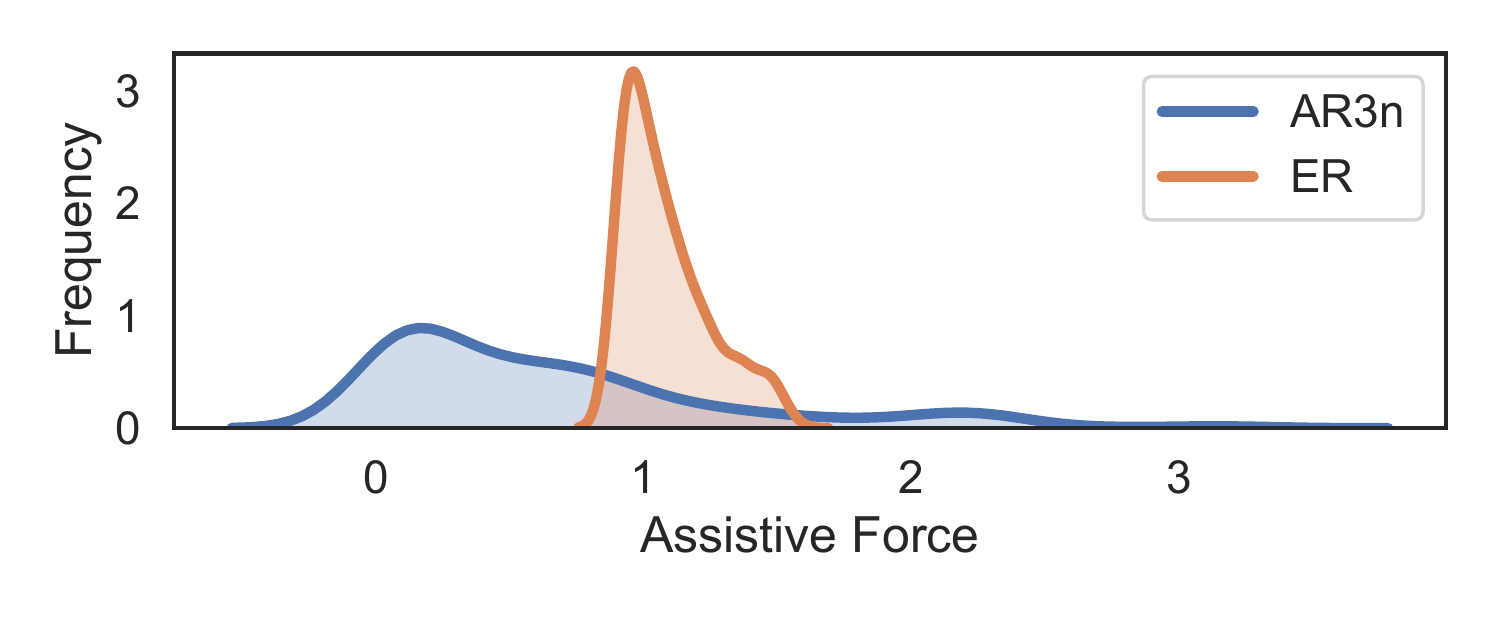}
	\caption{Distribution of assistive force under AR3n and ER.}
	\vspace{-10pt}
	\label{fig:force_real}
\end{figure}

Fig. \ref{fig:force_real} shows the distribution of assistive forces provided by AR3n and ER. ER's behavior was concentrated over a narrow region. The relatively \textit{narrow} distribution signifies the force field behavior of ER. In case of AR3n, assistive force was spread over a larger range. While AR3n mostly applied small corrective forces (higher density closer to zero), in some cases, it applied larger forces depending on the subject's performance. These observations reaffirm the ability of AR3n to provide assistance over multiple scenarios and highlight the inherent challenges of ER.

Next, we compared the performance of AR3n and ER as rehabilitative tools by comparing the change in tracking error between the baseline (T1) and final trial (T3) across the two assistance groups i.e. AR3n ($A_k$) and ER ($E_k$). The Shapiro-Wilk test was conducted on tracking error to verify normality at a p-value of $5\%$. All samples demonstrated normality and hence pairwise t-test were conducted.

\begin{figure}[!t]

	\centering
      \begin{subfigure}[t]{0.5\textwidth}
         \centering
    	\includegraphics[ width=\linewidth,trim={0 0cm 0 .5cm},clip]{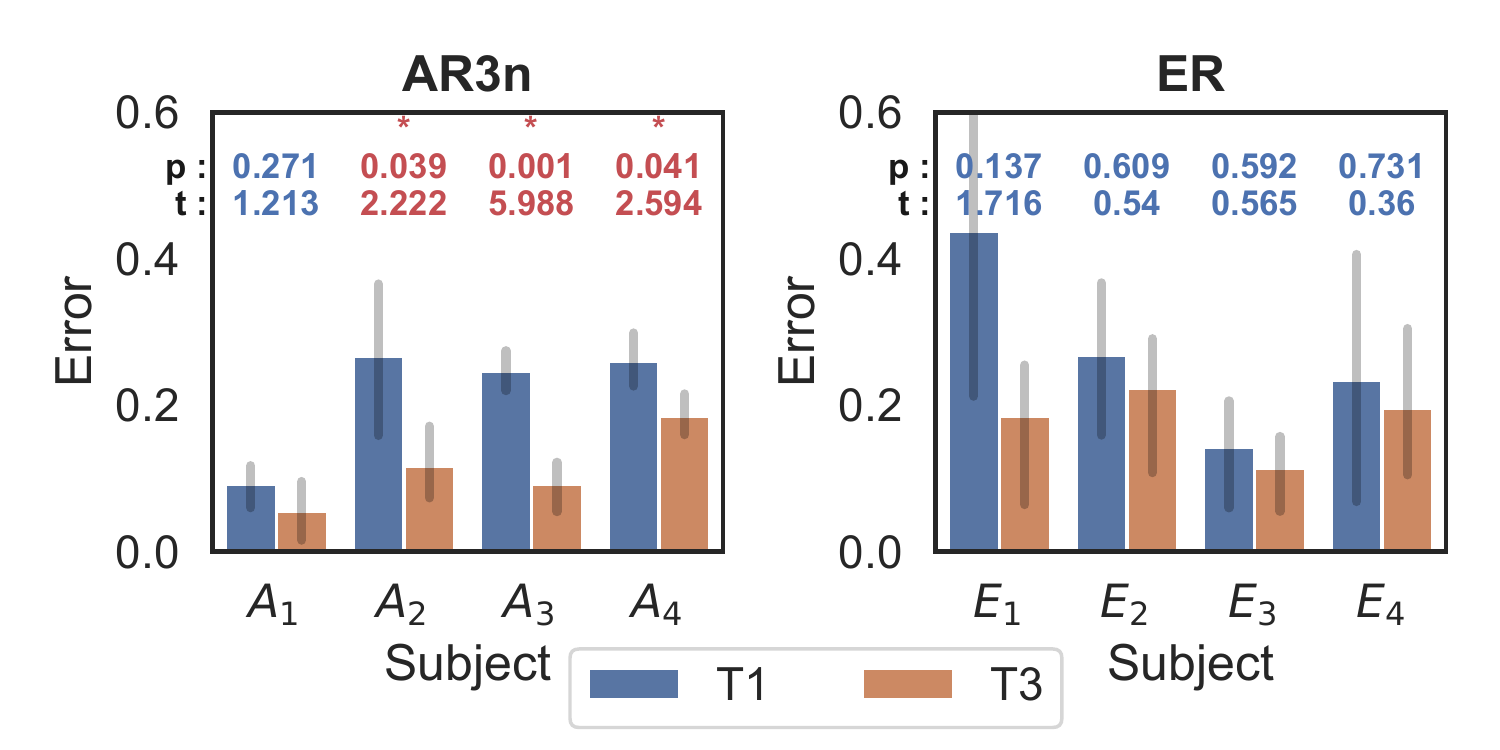}
     \end{subfigure}

     \begin{subfigure}[t]{0.5\textwidth}
         \centering
    	\includegraphics[ width=\linewidth,  trim={0 1cm 0 0cm},clip]{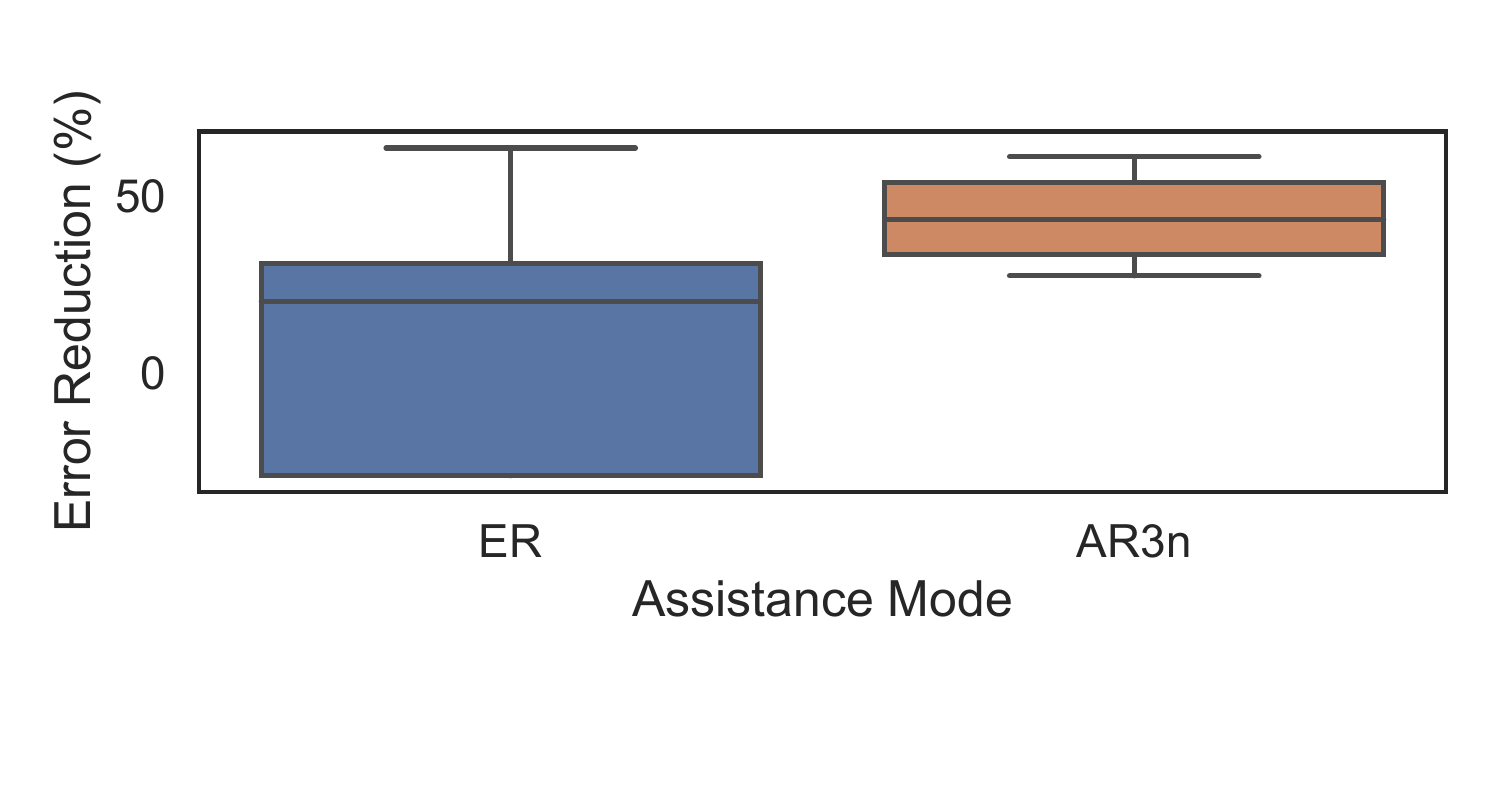}
     \end{subfigure}
	\vspace{-10pt}
 	\caption{(Top) Tracking error change between the baseline (T1) and final trial (T3) for test subjects. Text in red denotes significant changes in tracking error between trials. (Bottom) Box plots for percent change in tracking error between the first and last session under ER and AR3n assistance mechanism.}
   \label{fig:box_error}
\end{figure}

Fig. \ref{fig:box_error}-Top shows tracking errors for different subjects during T1 and T3. The blue bars denote the tracking error during the baseline recording (T1) while the red bars signify the final trial (T3). p-values and t-statistics obtained for pairwise t-tests on tracking error between T1 and T3 are also shown above the corresponding pairs. Pairs that demonstrated significant differences at $5\%$ are shown in red. p-values in blue denote that no significant differences were observed. 

None of the subjects in the ER group demonstrated significant reductions in tracking error over the duration of the experiment. The inferior performance from ER was expected and is attributed to the tendency of subjects to over-rely on robotic assistance, which led to a decline in performance when robotic assistance was removed. 

Three out of the four subjects under AR3n showed significant error reduction within the two trials. Only one subject ($A_1$) under AR3n that did not demonstrate significant error reduction. On closer inspection, $A_1$'s tracking error for T1 was lowest across the board when compared with all other subjects and trials, leaving very little scope for performance improvement over trials. Subjects under AR3n demonstrate a reduction in error variability as shown by the reduction in lengths of error bars from T1 to T3. On the other hand, subjects under ER display lower reduction in variance between sessions.

Finally, we also compared the percent error reduction across all subjects under AR3n and ER (see Fig. \ref{fig:box_error}-Bottom). Subjects in the AR3n group demonstrated higher improvements when compared to those within ER.

\section{CONCLUSION AND FUTURE DIRECTIONS} \label{sec:conclusion}

This paper describes a novel RL-based AAN controller called AR3n. AR3n uses SAC to modulate assistance in realtime based on a subject's performance. Using a reward function that minimizes tracking error while minimizing amount of assistive force enables the realization of a truly adaptive AAN controller. 

As opposed to traditional force field-based AAN controllers, AR3n does not require hand tuning of controller parameters. The system distinguishes itself from more sophisticated AAN controllers as our method does not require patient specific physical models. Instead, we simulate numerous virtual patients to generalize the controller over a larger population of subjects. The use of a virtual patient also distinguishes our method from previous RL-based AAN controllers \cite{obayashi2014assist, hamaya2016learning, zhangadaptive} that use online learning methods to generate subject-specific RL models.

We tested the proposed algorithm under numerous simulated and human subject experiments, and highlighted critical differences between AR3n and ER. AR3n demonstrated generalizability across multiple human subjects and efficacy as a rehabilitative tool. It was also observed that the method proposed here avoids over-reliance tendencies inherent of ER controllers.

Our system relies on offline learning to generate a subject-independent AAN controller. This method may not be suitable for patients with very specific needs. Use of online learning methods such as GARB \cite{weaver2013optimal} in conjunction with AR3n will enable the realization of controllers tuned to the specific needs of a patient without requiring extensive data collection. Future work should explore this option. 

Currently AR3n only modifies the gain of a proportional controller. It would be meaningful to design a study where derivative and integral gain values are modulated as well. Finally, the human subject study described in this paper was conducted on a fairly small pool of healthy subjects. Moving forward, the system needs to be tested with stroke patients and/or a larger subject pool to verify its scalability in the clinical setting.